\documentclass[conference]{IEEEtran}
\IEEEoverridecommandlockouts
\usepackage{cite}
\usepackage{amsmath,amssymb,amsfonts}
\usepackage{lipsum}
\usepackage{graphicx}
\usepackage{textcomp}
\usepackage{xcolor}
\usepackage{newunicodechar}
\newunicodechar{−}{\ensuremath{-}}
\usepackage{multirow}
\usepackage{breqn}
\usepackage{dirtytalk}
\usepackage{algorithm}
\usepackage{float}
\usepackage{graphicx}
\usepackage{upgreek}
\usepackage{algpseudocode}
\usepackage{booktabs}
\usepackage{subcaption}
\usepackage{graphicx}
\usepackage{multicol}
\makeatletter
\def\munderbar#1{\underline{\sbox\tw@{$#1$}\dp\tw@\z@\box\tw@}}
\makeatother

\ifCLASSOPTIONcompsoc
    \usepackage[caption=false, font=normalsize, labelfont=sf, textfont=sf]{subfig}
\else
\usepackage[caption=false, font=footnotesize]{subfig}
\fi
\def\BibTeX{{\rm B\kern-.05em{\sc i\kern-.025em b}\kern-.08em
    T\kern-.1667em\lower.7ex\hbox{E}\kern-.125emX}}
    
\usepackage{tikz}
\usepackage{url}
\usepackage{xcolor, colortbl}
\usetikzlibrary{decorations.pathreplacing}
\usetikzlibrary{arrows, shapes, positioning}
\usetikzlibrary{patterns}
\usepackage{pgfplots}
\usepackage{caption}
\pgfplotsset{compat=newest}
\pgfplotsset{plot coordinates/math parser=false}
\pgfkeys{/pgf/number format/.cd,1000 sep={\,}}
\usetikzlibrary{plotmarks}

\newlength\matlabfigurewidth
\newcolumntype{L}[1]{>{\raggedright\let\newline\\\arraybackslash\hspace{0pt}}m{#1}}
\newcolumntype{C}[1]{>{\centering\let\newline\\\arraybackslash\hspace{0pt}}m{#1}}
\newcolumntype{R}[1]{>{\raggedleft\let\newline\\\arraybackslash\hspace{0pt}}m{#1}}

\begin{document}

\title{VAE-GAN Based Price Manipulation in\\ Coordinated Local Energy Markets \vspace{-2mm}
}


\author{\IEEEauthorblockN{\textbf{Biswarup Mukherjee}\IEEEauthorrefmark{1}, \textbf{Li Zhou}\IEEEauthorrefmark{1}, \textbf{S. Gokul Krishnan}\IEEEauthorrefmark{1}, \textbf{Milad Kabirifar}\IEEEauthorrefmark{2}, \\ \textbf{Subhash Lakshminarayana}\IEEEauthorrefmark{2}, \textbf{Charalambos Konstantinou}\IEEEauthorrefmark{1}
}
\IEEEauthorblockA{
\IEEEauthorrefmark{1}CEMSE Division, King Abdullah University of Science and Technology (KAUST), Saudi Arabia\\
\IEEEauthorrefmark{2}School of Engineering, University of Warwick, Coventry, United Kingdom 
}

\vspace{-2.6em}}

\maketitle
\IEEEpubidadjcol

\begin{abstract}

This paper introduces a model for coordinating prosumers with heterogeneous {distributed energy resources (DERs)}, participating in the local energy market (LEM) that interacts with the market-clearing entity. {The proposed LEM scheme utilizes a data-driven, model-free reinforcement learning approach based on the multi-agent deep deterministic policy gradient (MADDPG)} framework, enabling prosumers to make real-time decisions on whether to buy, sell, or refrain from any action while facilitating efficient coordination for optimal energy trading in a dynamic market. {In addition, we investigate a price manipulation strategy using a} variational auto encoder-generative adversarial network (VAE-GAN) model, which allows utilities to adjust price signals in a way that induces financial losses for the prosumers. Our results show that under adversarial pricing, heterogeneous prosumer groups, particularly those lacking generation capabilities, incur financial losses. The same outcome holds across LEMs of different sizes. As the market size increases, trading stabilizes and fairness improves through emergent cooperation among agents. 
\end{abstract}
\vspace{-4mm}
\section{Introduction} 
The modern electric grid is shifting from a centralized structure toward greater use of \textit{distributed energy resources} (DERs), driven by the growing adoption of renewable technologies like \textit{photovoltaic} (PV) panels and wind turbines installed near end users. However, the intermittent nature of renewables introduces variability and uncertainty, posing challenges to grid stability. To address these, effective DER management is essential. \textit{Local energy markets} (LEMs) have emerged as a promising solution for coordinating energy transactions among participants \cite{pinto2021local, TSAOUSOGLOU2022111890}. By enabling direct energy trading between consumers and prosumers, LEMs enhance system flexibility, support renewable integration, and reduce costs. They also alleviate stress on distribution networks, improve reliability, and optimize infrastructure use, potentially reducing the need for costly upgrades \cite{morstyn2018using, qiu2021scalable}.

In recent years, various approaches have been proposed to coordinate participants in LEMs \cite{8501599, MAY2023120705, 9133518, 2024_LEMCordWily}. These fall into two main categories: (a) model-based methods, which use predefined mathematical models to optimize energy trading, and (b) model-free methods, which rely on data-driven techniques like \textit{reinforcement learning} (RL) to learn optimal strategies directly from market interactions. In deregulated electricity markets, traditional bidding involves generation companies (GENCOs) submitting capacity and pricing offers, with an \textit{independent system operator} (ISO) determining the \textit{market-clearing price} (MCP) based on bids and system constraints \cite{Zhang2015}. The growing share of renewables has led to structural changes in market design, including: (i) shorter trading intervals to manage variability \cite{KOCH2019109275}; (ii) broader participation from microgrids, \textit{virtual power plants} (VPPs), and demand-response systems \cite{PANOS2019104476}; and (iii) tighter integration between energy and ancillary service markets \cite{8957676}.

With the evolving market landscape, effective bidding strategies must account for the strategic interactions among participants, where each agent’s outcome depends on the actions of others. This interdependence underscores the importance of estimating a \textit{Nash Equilibrium} (NE), in which no participant benefits from unilaterally deviating \cite{10.1007_91}. Traditional methods such as risk-averse stochastic optimization \cite{8558521}, robust optimization \cite{pousinho2015robust}, and distributionally robust optimization \cite{han2020distributionally} often neglect the dynamic nature of competitors’ bid adjustments. To overcome this limitation, \textit{multi-agent reinforcement learning} (MARL) has gained traction as it captures real-time strategic adaptation and coordination in competitive bidding environments \cite{SAMENDE2022119123, MAY2023120705, 9133518, 8496766}. Specifically, in LEMs, where prosumers can trade surplus energy, RL has been applied to optimize trading strategies without relying on predefined market equations. RL agents learn optimal actions, buying, selling, or remaining inactive, through interaction with the market environment. For example, deep Q-learning has been used to model prosumer behavior \cite{8496766}, though it faces scalability challenges. To address this, \textit{multi-agent deep deterministic policy gradient} (MADDPG) offers a more scalable and flexible framework for learning competitive trading strategies in multi-agent settings \cite{SAMENDE2022119123}.

Both market clearing and bidding optimization exhibit the Markov property, where decisions depend on prior system states. Traditional optimization methods often struggle to adapt to evolving market dynamics, while deep RL can learn effective decision-making strategies from historical interactions. Moreover, unlike centralized equilibrium computations, deep RL
supports distributed and decentralized learning, well-suited for competitive environments where participants cannot share strategies due to privacy concerns \cite{9705504,9133518}. Deep RL is therefore promising for: (a) bidding strategies, where prosumers and GENCOs dynamically adjust bids based on past market-clearing prices to approximate equilibrium behavior; and (b) dispatch strategies, where ISOs optimize energy allocation in real time, reducing the risk of market manipulation.

{An often-overlooked challenge in LEMs is price signal manipulation, where \textit{distribution system operators} (DSOs)  can adjust price signals to influence prosumer decisions. Traditionally, system operators monitor market conditions to prevent price manipulation and market power abuse \cite{9351299}, but advanced adversarial strategies can bypass these detection mechanisms \cite{6074981}. In this context, RL-based methods can be used to simulate price manipulation and develop countermeasures. Recent studies have shown that adversarial learning techniques can generate fake market signals that distort prosumers’ decision-making processes \cite{wang2020market}. Our study extends this by integrating a hybrid  \textit{variational autoencoder-generative adversarial network} (VAE-GAN) model to explore how utilities can manipulate price signals to incentivize prosumers to sell energy at lower prices, consolidating utility dominance in the market. GANs have been widely used for years as powerful artificial \textit{neural network} (NN)-based models for generating synthetic data \cite{10.1145/3422622, 9578137}. However, as reported in \cite{SCHULTZ2024110138}, GANs often struggle with small tabular datasets.  In contrast, the hybrid VAE-GAN mitigates this challenge by leveraging the VAE’s ability to encode complex latent structures, allowing it to extract meaningful features from limited data. This enhances generalization compared to traditional GAN-based models.
Moreover, VAE-GAN is inherently more resilient to noise, ensuring that the generated data maintains variability without overfitting to the training samples~\cite{HSLVAEGAN}.} 

{The contributions of this work are as follows:\\
\noindent (1) We develop a MARL-based framework utilizing the MADDPG algorithm \cite{SAMENDE2022119123} to model prosumer trading in dynamic bidding environments. The approach enables prosumers to autonomously adapt strategies through ongoing market interaction, enhancing responsiveness to price fluctuations. \\
\noindent (2) We introduce a hybrid VAE-GAN model to analyze adversarial price manipulation, demonstrating how utilities and system operators can distort signals to influence prosumer behavior and expose vulnerabilities in decentralized trading. \\
\noindent (3) Departing from NE-based market-clearing models, we explore prosumer responses under adversarial market conditions, offering insights into how diverse agents adapt bidding strategies in dynamic and potentially exploitative settings.

The remainder of the paper is organized as follows. Section II presents the RL-based energy trading framework for multi-agent prosumers in LEMs. Section III introduces the VAE-GAN-based price manipulation model. Section IV details the MADDPG-based trading strategy. Sections V and VI cover the case study and results, and Section VII concludes the paper.

\section{Energy Trading for Prosumers using Multi-Agent RL Framework}
In a multi-agent system, each agent's actions influence both the environment and other agents, forming a Markov Decision Process (MDP) \( M = (S, A, \tau, R, \gamma) \), where:
\begin{itemize}
    \item \( S = \{S_1, \ldots, S_n\} \): set of states representing the agent and its environment.
    \item \( A = \{A_1, \ldots, A_k\} \): set of possible actions.
    \item \( \tau(s, a, s') \): transition function defining the probability of moving from state \( s \) to \( s' \) after taking action \( a \).
    \item \( R \): reward function providing feedback for each action.
    \item \( \gamma \): discount factor for future rewards.
\end{itemize}

\begin{figure}[t]
    \centering
    \begin{minipage}{0.5\textwidth} 
        \centering
        \includegraphics[width=0.65\textwidth]{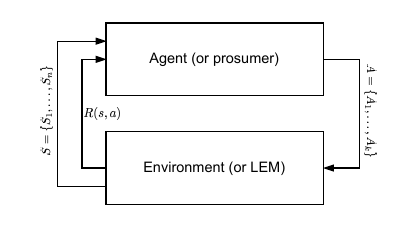} 
        \vspace{-3mm}
        \caption{Illustration of prosumer interaction with the LEM at time interval $t$.}
        \vspace{-4mm}
        \label{fig:example}
    \end{minipage}
\end{figure}

A prosumer participating in a LEM observes the \textbf{state} \( s^t \) at each time step \( t \), which includes: (a) battery state-of-charge (SOC), indicating stored energy; (b) energy demand; (c) power generation; and (d) market price. Based on this, the prosumer selects an \textbf{action} \( a^t \): buying energy if demand exceeds supply, selling surplus if generation exceeds consumption, or taking no action. The environment then transitions to a new state \( s^{t+1} \), adjusting battery levels, financial outcomes, and market conditions. The prosumer receives a \textbf{reward} \( R(s, a) \), such as profit or cost. This iterative process enables learning an optimal policy \( \pi: S \to A \) to maximize long-term rewards.

\subsection{Multi-Prosumer Modeling for LEM}
In a MARL setup, each prosumer \( i \) uses a policy network (MLP) to select actions based on its state and maximize cumulative rewards in a simulated market. The \texttt{state space} \( S \) and \texttt{action space} \( A \) are defined as:
\begin{align}
S &= \left\{ s_i^{t+1} \mid s_i^t \right\} = \left\{d_i^t, g_i^t, \text{SOC}^0_{i}, \hat{\lambda}^t \right\} \\
A &= a_i^t = \left\{\text{Buy}, \text{Sell}, \text{No-Op} \right\}
\end{align}
where \( d_i^t \), \( g_i^t \), and \( \hat{\lambda}^t \) are the demand, PV generation, and market price at time \( t \); \( \text{SOC}^0_i \) is the initial battery state-of-charge. State transitions follow \( \tau(s^{t+1} \mid s^t, a^t) \).

Each prosumer must satisfy a power balance:
\begin{align}
\hat{e}_i^{t} = d_i^t + p^{\downarrow,t}_i - \left( p^{\uparrow,t}_i + g_i^t \right)
\end{align}
and update its SOC as:
\begin{align}
\text{SOC}_i^t = \text{SOC}_i^0 + \frac{\Delta T}{E^{\max}_i} \left( \eta^\uparrow p_i^{\uparrow,t} - \frac{p_i^{\downarrow,t}}{\eta^\downarrow} \right)
\end{align}
with bounds \( \overline{\text{SOC}} \le \text{SOC}_i^t \le \underline{\text{SOC}} \), and where \( \eta^\uparrow, \eta^\downarrow \) are charging/discharging efficiencies.

Unmet demand and surplus are:
\begin{subequations}
\begin{align}
e_i^{d,t} &= \max\left\{0, \hat{e}_i^t\right\} \\
e_i^{s,t} &= \min\left\{0, \hat{e}_i^t\right\}
\end{align}
\end{subequations}

Charging/discharging powers are capped: \( p_i^{\uparrow,t}, p_i^{\downarrow,t} \le \Bar{u} \). The purchasable power is:
\begin{align}
p_i^{\uparrow,t} = 
\begin{cases} 
0 & \text{if } |e_i^{s,t}| \ge \min(\underline{u}, \Bar{u}) \\
\min(\underline{u}, \Bar{u}) - |e_i^{s,t}| & \text{if } |e_i^{s,t}| < \min(\underline{u}, \Bar{u}) \\
e_i^{d,t} + \min(\underline{u}, \Bar{u}) & \text{if } e_i^{d,t} \ge 0
\end{cases}
\end{align}
where \( \underline{u} = \frac{E_i^{\max}}{\Delta T} \).%
\footnote{We assume \( \Bar{u} \ge \underline{u} \) for all \( i \).}

\subsection{Market Equilibrium}
From a market equilibrium perspective, the clearing entity balances supply and demand by allocating energy among prosumers who follow learned policies in a MARL framework. At any time \( t \), let \( m \) prosumers possess surplus energy and \( k \) prosumers experience a shortfall. The total available surplus \( S^t \) and the total demand deficit \( D^t \) are computed as:
\begin{subequations}
\begin{align}
& S^t = \sum_{m} \left|e_i^{s,t}\right|, && {\text{$\forall t$}} \\
& D^t = \sum_{k} e_i^{d,t}, && {\text{$\forall t$}}.
\end{align}
\end{subequations}
When \( D^t \geq S^t \), the surplus is distributed proportionally, with the remaining demand being supplied by the grid:
\begin{subequations}\label{eq:higherDemand_clearing}
\begin{align}
& \sum_m \left|e_i^{s,t}\right| = \sum_k \delta D^t \cdot e_i^{d,t}, && {\text{$\forall i ~\text{and}~ t$}} \\
& e_i^{g,t} = (1 - \delta D^t) \cdot e_i^{d,t}, && {\text{$\forall i ~\text{and}~ t$}}
\end{align}
\end{subequations}
where \( \delta D^t = \frac{S^t}{D^t} \) defines the fraction of demand allocated from surplus prosumers. When \( D^t < S^t \), power is sold to the grid:
\begin{subequations}
\begin{align}
& \delta S^t = \frac{1}{\delta D^t}, && \forall t \\
& \sum_k e_i^{d,t} = \sum_m \delta S^t \cdot \left|e_i^{s,t}\right|, && {\text{$\forall i ~\text{and}~ t$}} \\
& e_i^{g',t} = (1 - \delta S^t) \cdot \left|e_i^{s,t}\right|, && {\text{$\forall i ~\text{and}~ t$}}.
\end{align}
\end{subequations}
This proportional allocation prioritizes local energy consumption before purchasing from or selling to the grid.

In addition to the above, several requirements are imposed from the perspective to maintain market balance. First, agents cannot purchase energy if storage or generation is available, ensuring energy is only bought when necessary. Second, agents may only sell energy during the morning when market prices are higher and PV generation is active, encouraging the use of favorable market conditions. Third, agents can buy energy during post-evening and early morning hours (e.g., 5 PM to 8 AM), helping to manage peak load conditions and support grid stability.

\section{Hybrid VAE-GAN based Price Manipulation}
This section introduces the hybrid VAE-GAN for adversarial price manipulation, causing financial losses for prosumers in the LEM. As automated pricing becomes more common, we explore how malicious actors, like DSOs or market participants, could manipulate market outcomes. We first provide background on VAE-GAN and its components before presenting the mathematical formulation.

\subsection{Overview of VAE-GAN}
A VAE-GAN combines a \textit{variational autoencoder} and a \textit{generative adversarial network} to generate realistic data distributions while preserving latent representations. The encoder (\(E\)) compresses input data (e.g., price signals) into a lower-dimensional latent space capturing key statistical features. The decoder (\(G\)) reconstructs signals from this space, resembling historical trends, while the discriminator (\(D\)) distinguishes real from generated data, improving output realism. The VAE-GAN in this work reconstructs electricity price signals and generates manipulated versions that appear realistic but serve adversarial goals. By learning the underlying price distribution, the NN model can strategically adjust these signals to induce financial losses for prosumers.

\subsection{The Price Manipulation Strategy}
A key part of our approach is manipulating the \textit{real-time price} (RTP), which is used for utility-prosumer transactions. Unlike traditional markets with a ``clearing price'', the LEM clearing mechanism relies on pro-rata energy sharing, meaning there is no explicit ``clearing price''. Instead, the utility dynamically adjusts the RTP, raising it when a prosumer buys energy and lowering it when they sell, as described in Algorithm \ref{alg:VAE-GAN}. The manipulated price signal is adjusted such that: (1) The price falls below the RTP when the prosumer buys energy, causing them to pay more than under a fair transaction; (2) The price rises above the RTP when the prosumer sells energy, reducing their earnings. 
The VAE-GAN outputs the manipulated price signal, incorporating a deviation term $\delta\lambda^t$. The reconstructed price signals are:
\begin{subequations}
\begin{align}
& \tilde{\lambda}^{t,\downarrow}_i = \min \left\{\hat{\lambda}^t, \hat{\lambda}^{gan} +\delta{\lambda}^{t} \right\},  \label{eq:price_low} && {\text{$\forall i ~\text{and}~ t$}}  \\
&  \tilde{\lambda}^{t,\uparrow}_i = \max \left\{\hat{\lambda}^t, \hat{\lambda}^{gan} +\delta{\lambda}^{t} \right\}, \label{eq:price_up} && {\text{$\forall i ~\text{and}~ t$}}.
\end{align}
\end{subequations}
Here, $\hat{\lambda}^{gan}$ represents the reconstructed price signal generated by the GAN, and $\delta\lambda^t$ is drawn from a time-varying distribution, $\delta\lambda^t \sim \mathcal{N}(\mu_t, \sigma_t^2)$, which enhances the model's robustness against noise and improves data quality~\cite{zhong2020generative}.

\begin{algorithm}[b!]
    \caption{VAE-GAN based price manipulation} 
    \label{alg:VAE-GAN}
    \begin{algorithmic}[1]
        \State \textbf{Initialize:} Encoder $E$, Decoder $G$ \& Discriminator $D$
        \State \textbf{Set} hyperparameters: learning rates and epochs
        \For{each epoch}

            \State Take $\hat{\lambda}$ as input
            \State Compute energy balance 
            \State \textbf{VAE Training:}
            \State $\mu, \sigma \gets \text{E}(\hat{\lambda}) $
            \State $z \gets \mu + \sigma \cdot \epsilon; \epsilon \in \mathcal{N}(0, 1)$  
            \State $\hat{\lambda}' \gets G(z)$
            \State Compute $\mathcal{L}_{vae}$ using \eqref{eq:vae_loss}
            
            \State \textbf{Discriminator Training:}
            \State \textbf{Compute} $\mathcal{L}_{D}$ using \eqref{eq:generator_loss}
            \State \textbf{Energy balance based price manipulation:}
            \For{each time step \( t \) in batch}
                \If{$ e_i^{s,t} \le 0 $ }
                    \State Reduce price $\tilde{\lambda}^{t,\downarrow}_i$ using \eqref{eq:price_low}
                \Else
                    \State Increase price $\tilde{\lambda}^{t,\uparrow}_i$ using \eqref{eq:price_up}
                \EndIf
            \EndFor
            \State \textbf{Compute} $\mathcal{L}_{total}$
            \State \textbf{Update} Parameters back-propagating $\mathcal{L}_{total}$

        \EndFor
    \State Output: $\tilde{\lambda}^{\downarrow}_i$, $\tilde{\lambda}^{\uparrow}_i$
    \end{algorithmic}
\end{algorithm}

\subsection{VAE-GAN Training and Loss Function Model}
The VAE-GAN training follows a standard adversarial approach for price manipulation (Algorithm~\ref{alg:VAE-GAN}). The VAE reconstructs the price signal based on a compressed latent space representation of the input RTP data ($\hat{\lambda}^t$). The VAE learns the distribution of the price signals, enabling it to detect anomalies by comparing the original signal to its reconstruction. In the VAE-GAN, a discriminator \textit{D} competes with the decoder \textit{G} to generate a realistic price signal $\hat{\lambda}'$ that closely matches the original distribution. The training process computes the loss for each epoch, then updates the parameters using backpropagation. The VAE loss is computed as:
\begin{align}
\mathcal{L}_{vae} = \text{MSE}(\hat{\lambda}', \hat{\lambda}) + D_{KL}[q(z|\hat{\lambda}') || p(z)] \label{eq:vae_loss}
\end{align}
For the discriminator, the loss is:
\begin{align}
\mathcal{L}_{D} = -\mathbb{E}[\log D(\hat{\lambda})] + \mathbb{E}[\log(1 - D(G(z)))] \label{eq:generator_loss}
\end{align}
where $z$ is the encoded price signal and $\hat{\lambda}'$ is the decoded signal. The total VAE-GAN loss is:
\begin{align}
\mathcal{L}_{total} = \mathcal{L}_{vae} + \zeta \left(\mathcal{L}_{D}\right) \label{eq:total_loss}
\end{align}

A key assumption in our analysis is that the utility (or DSO) has full knowledge of each prosumer’s demand, generation, and storage policy (see line 5, Algorithm~\ref{alg:VAE-GAN}). This depends on regulatory, privacy, and metering infrastructure factors.

\vspace{-1.0mm}
\section{Optimal Energy Trading for LEM}

\subsection{LEM Modeling Using Actor-Critic Method}

To model the LEM, we adopt the MADDPG method, that leverages deep neural networks for both the policy (actor) and the value function (critic) \cite{2024_LEMCordWily}. Each agent comprises: (a) an \textit{actor} that selects optimal actions based on the policy, and (b) a \textit{critic} that evaluates these actions by estimating expected $Q$-values to guide actor improvement. Market balancing and other constraints are embedded within the MADDPG framework to govern energy transactions and support efficient grid management, effectively capturing the LEM's dynamics.

\vspace{-1mm}
\subsection{Modeling Rewards for Prosumers}
We model prosumer rewards by defining a utility function that captures both current and future gains. As in \cite{MAY2023120705}, the utility at time $t$ considers income from selling $e_i^{s,t}$, costs from buying $e_i^{d,t}$, and a risk factor $\eta$:
\begin{align}
u^t_i = \frac{\left\{|e_i^{s,t}| \cdot \hat{\lambda}^t\right\}^{(1 - \eta)} - 1}{1 - \eta} - \left\{e_i^{d,t} \cdot \hat{\lambda}^t \right\}.
\end{align}
The immediate reward for each prosumer is then:
\begin{align}\label{eq:reward}
r^t_i = 
\begin{cases} 
u^t_i - u^{t-1}_i & \text{if Sell} \\ 
-(u^t_i - u^{t-1}_i) & \text{if Buy} \\ 
0 & \text{if No-op.}
\end{cases}
\end{align}
The reward function incentivizes selling (positive reward) and penalizes buying (negative reward), motivating prosumers to maximize cumulative rewards. The $Q$-value, based on the Bellman equation, uses the immediate reward and discounted future rewards (factor $\gamma$) to guide policy updates.

\vspace{-1mm}
\subsection{Training and Execution for MLP}
For each agent \(i \in I\), the critic networks are parameterized by \(\phi_i\), with losses defined as:
\begin{subequations}
\begin{align}
\mathcal{L}(\phi_i) = \mathbb{E} \left[ \left( y_i^t - Q_i(s_i^t | \phi_i) \right)^2 \right], && \forall i \label{eq:attack_critic_loss}
\end{align}
where $y_i$ denotes the target Q-value. The corresponding actor losses, with \(\theta_i\) denoting the actor parameters, are:
\begin{align}
\mathcal{L}(\theta_i) = -\mathbb{E}_{s_i^t} \left[ Q_i(s_i^t, \pi_i(s_i^t | \theta_i)) \right], && \forall i. \label{eq:joint_actor_loss}
\end{align}
\end{subequations}
Actor updates follow the deterministic policy gradient:
\begin{align}\label{eq:Policy_grad}
\nabla J = \mathbb{E} \left[ \nabla_a Q_i(s_i^t, \cdot | \phi_i) \big|_{a_i^t = \pi_i} \cdot \nabla_{\theta_i} \pi_i(\cdot | \theta_i) \right].
\end{align}
Gradients guide parameter updates to minimize loss and improve action selection. The Adam optimizer adjusts the actor’s policy accordingly. The full MADDPG-based training and execution strategy is detailed in Algorithm~\ref{alg:maddpg_dual}.

\begin{algorithm}[h!]
\caption{MADDPG training and execution with actors and critics}
\label{alg:maddpg_dual}
\begin{algorithmic}[1]
\State \textbf{Initialize} parameters $\phi_i$ of critic for each $i \in I $ 
\State \textbf{Initialize} parameters $\theta_i$ of actor for each $i \in I$ 
\For{$e: epoch = 0, 1, 2, \dots$}
\For{$t: episode = 0, 1, 2, \dots$}
    \State \textbf{Observe} States $[s_1^t, s_2^t, \dots, s_n^t]$
        \For{each agent $i \in I$}
            \State $a_i^{t} \gets \pi_i(s_i^t | \theta_i$)
        \EndFor
\EndFor    
\State \textbf{Apply actions}
\State \textbf{Collect new observations and rewards}
    
\For{each agent $i \in I$}
        \If {$s_{t+1}$ is terminal}
            \State $Q_i^t = r_i^{t}$ 
        \Else
                \State $Q_i^t \gets r_i^{t} + \gamma Q_i^{t+1}(s_i^{t+1}, a' | \phi_i)$
        \EndIf
            \State \textbf{Calculate actor loss} $\mathcal{L}(\theta_i) $ using \eqref{eq:joint_actor_loss}
            \State \textbf{Calculate critic loss} $\mathcal{L}(\phi_i)$ using \eqref{eq:attack_critic_loss}
            \State \textbf{Update} parameters $\theta_i$ to minimize $\mathcal{L}(\theta_i)$
            \State \textbf{Update} parameters $\phi_i$ to minimize $\mathcal{L}(\phi_i)$
\EndFor
\State \textit{Save actor $\theta_i$ and critic  critic $\phi_i$ networks periodically}
\EndFor \\
\textbf{Output}: Best actor $\theta_i$ and critic $\phi_i$
\end{algorithmic}
\end{algorithm}

\vspace{-3mm}
\section{Case Study}

\begin{table*}[t]
\centering
\vspace{1mm}
\caption{Prosumer and their resource combinations in different LEMs.}\label{tab1}
\vspace{-2mm}
\begin{tabular}{>{\centering\arraybackslash} p{0.15\textwidth}||>{\centering\arraybackslash} p{0.18\textwidth} || >{\centering\arraybackslash} p{0.18\textwidth} || >{\centering\arraybackslash} p{0.3\textwidth}}
\hline
\hline
\textbf{Group} & \textbf{LEM A Prosumer ID} & \textbf{LEM B Prosumer ID} & \textbf{Resource Combinations} \\
\hline
\hline
Group 1 & 1 - 8  & 1 - 38 & No PV and No ESS \\
\hline
Group 2 & 9 - 13 & 39 - 63 & No PV + ESS \\
\hline
Group 3 & 14 - 16 & 64 - 76 & PV + No ESS \\
\hline
Group 4 & 17 - 20 & 77 - 100 & PV + ESS \\
\hline
\hline
\end{tabular}
\vspace{-4mm}
\end{table*}

For the case study, we examine two heterogeneous LEMs: a smaller one with 20 prosumers (LEM A) and a larger one with 100 prosumers (LEM B). The resource configurations for these groups are listed in Table~\ref{tab1}. Fig.~\ref{fig:av_demand_all_group} shows the average demand profiles for prosumers in both LEMs, with higher consumption during morning and evening hours. PV generation for Groups 3 and 4 is depicted in Fig.~\ref{fig:pv_average}. To emulate real-world variability, Gaussian noise \(\mathcal{N}(0, 0.05)\) is added to the demand, following the methodology in \cite{Ratnam14092017}. The 24-hour demand and PV generation data are extended to 48 hours for simulation. All prosumers are assumed to have identical energy storage capacity (\(E_{max}\)) with either 5 kWh or 13 kWh batteries. Storage efficiency, both charging (\(\eta^\uparrow\)) and discharging (\(\eta^\downarrow\)), is uniformly set to 0.7. The price signal \(\hat{\lambda}^t\), sourced from a spot market \cite{electricityPrice}, is shown in Fig.~\ref{fig:pv_average}. We use a risk factor \(\eta = 0.5\) in the utility function. Hyperparameters for the neural network algorithm are detailed in Table~\ref{tab:hyper}.

\begin{figure}[t]
    \centering
    \begin{minipage}{0.4\textwidth} 
        \centering
        \includegraphics[width=\columnwidth]{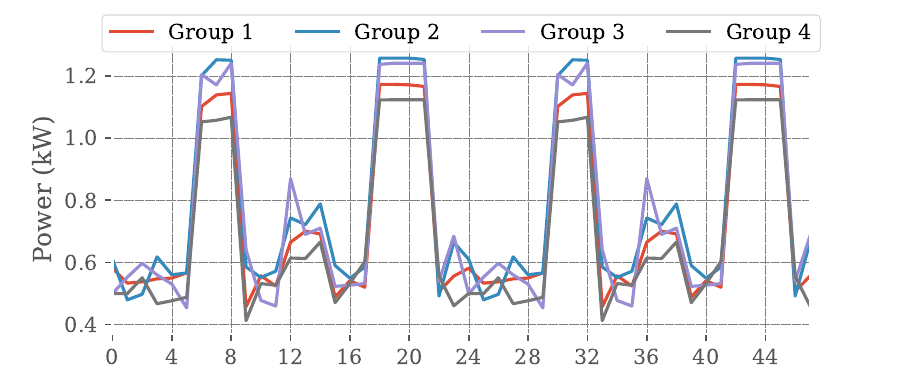} 
        \vspace{-5mm}
        \caption{Average demand profile for prosumers of four groups.}
        \label{fig:av_demand_all_group}
    \end{minipage}
    \hfill
    \begin{minipage}{0.4\textwidth} 
        \centering
        \includegraphics[width=\columnwidth]{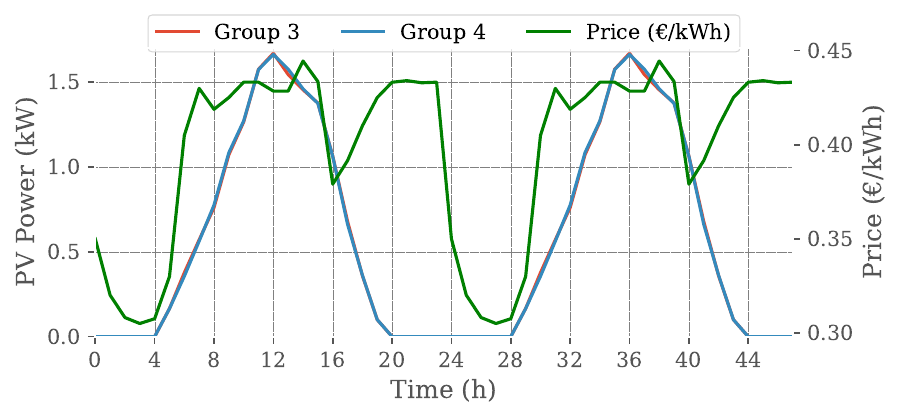}
        \vspace{-5mm}
        \caption{Average PV generation profile for prosumers of two different Groups in the LEM and the input spot market price.}
        \label{fig:pv_average}
    \end{minipage}
    \vspace{-5mm}
\end{figure}

\begin{table}[h!]
\caption{NN hyper-parameters and their values.}
\vspace{-2mm}
\label{tab:hyper}
\centering
{\fontsize{8}{8}
\begin{tabular}{>{\centering\arraybackslash} p{0.28\textwidth} || >{\centering\arraybackslash} p{0.15\textwidth}}
\hline
\hline
\textbf{Hyper-parameter details}     & \textbf{Values} \\
\hline
\hline
Epochs      & 100            \\ \hline
Episodes Per Epoch     & 48            \\ \hline
MADDPG-Actor Layer 1 size                         & 9              \\ \hline
MADDPG-Actor Layer 2 size     & 64   \\ \hline
MADDPG-Actor Layer 3 size     & 3    \\ \hline
MADDPG-Critic Layer 1 size & 240 \\ \hline
MADDPG-Critic Layer 2 size & 64 \\ \hline
MADDPG-Critic Layer 3 size & 64 \\ \hline
MADDPG-Critic Layer 4 size & 1 \\ \hline
MADDPG-Actor Learning rate ($\alpha$)  & 0.01    \\ \hline
MADDPG-Critic Learning rate ($\beta$) & 0.1 \\ \hline
Discount factor ($\gamma$) & 0.95 \\ \hline
Mean ($\mu_t$) & 0.4 \\ \hline
Standard deviation ($\sigma_t$) & 0.1 \\ \hline
Balance factor ($\zeta$) & 1 \\ \hline
VAEGAN-Encoder Layer 1 size  & 24 \\ \hline
VAEGAN-Encoder Layer 2 size & 32 \\ \hline
VAEGAN-Encoder Layer 3 Size & 16 \\ \hline
VAEGAN-latent Size & 24 \\ \hline
VAEGAN-Decoder Layer 1 size & 24 \\ \hline
VAEGAN-Decoder Layer 2 size & 16 \\ \hline
VAEGAN-Decoder Layer 3 size & 32 \\ \hline
VAEGAN-Decoder Layer 4 size & 24 \\ \hline
VAEGAN-Discriminator Layer 1 size & 24 \\ \hline
VAEGAN-Discriminator Layer 2 size & 32 \\ \hline
VAEGAN-Discriminator Layer 3 size & 16 \\ \hline
VAEGAN-Discriminator Layer 4 size & 1 \\ \hline
VAEGAN-Learning Rate & 0.001 \\ 
\hline
\hline
\end{tabular}}
\vspace{-4mm}
\end{table}

\vspace{-2mm}
\section{Results and Discussion}
\vspace{-1mm}

\begin{figure}[t] 
    \centering
    \begin{minipage}{0.4\textwidth} 
        \centering
        \includegraphics[width=\textwidth]{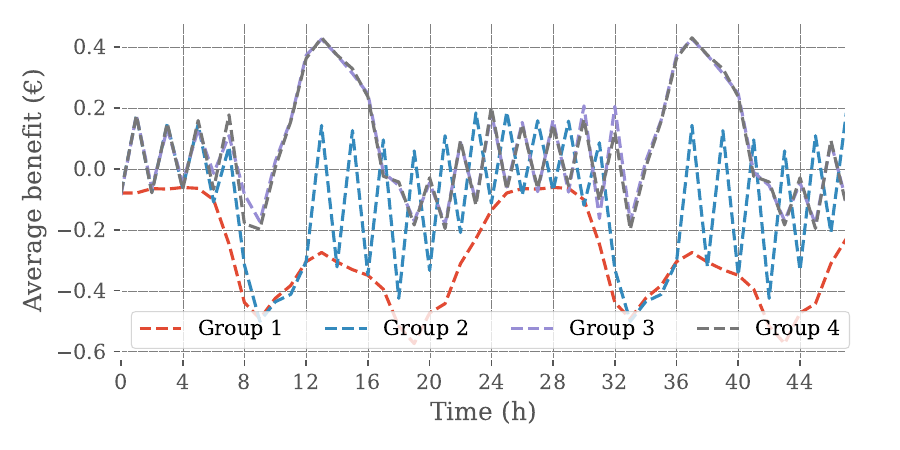} 
        \vspace{-6mm}
        \subcaption{Without price-manipulation in LEM A.} \label{fig:prosumer_avg_benefits_1} 
    \end{minipage}
    \hfill
    \begin{minipage}{0.4\textwidth} 
        \centering
        \includegraphics[width=\textwidth]{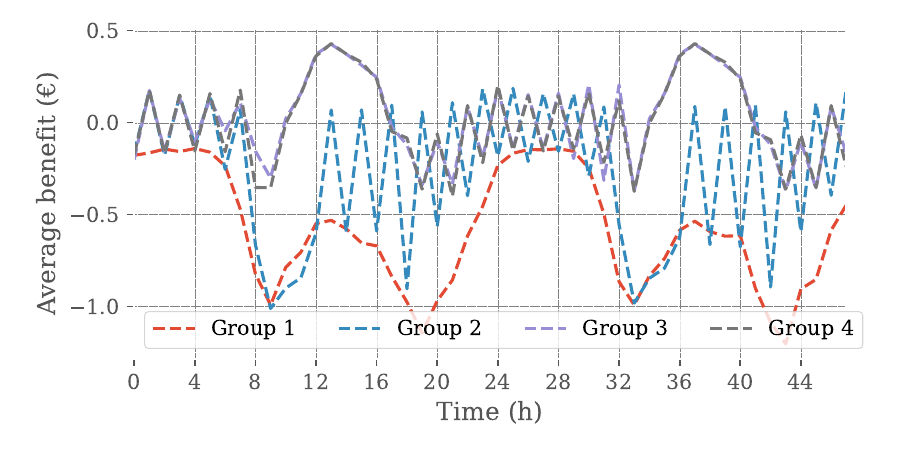} 
        \vspace{-6mm}
        \subcaption{With price-manipulation in LEM A.} \label{fig:prosumer_avg_benefits_2} 
    \end{minipage}
    \begin{minipage}{0.4\textwidth} 
        \centering
        \includegraphics[width=\textwidth]{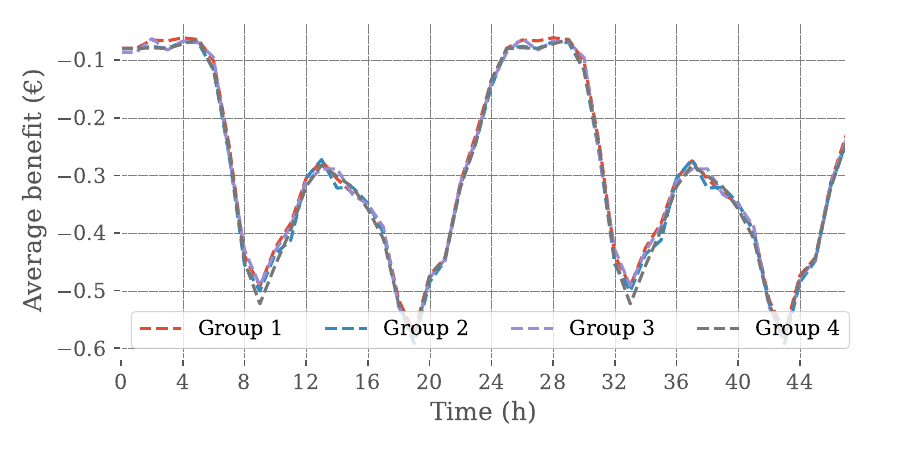} 
        \vspace{-6mm}
        \subcaption{Without price-manipulation in LEM B.} \label{fig:prosumer_avg_benefits_3} 
    \end{minipage}
    \begin{minipage}{0.4\textwidth} 
        \centering
        \includegraphics[width=\textwidth]{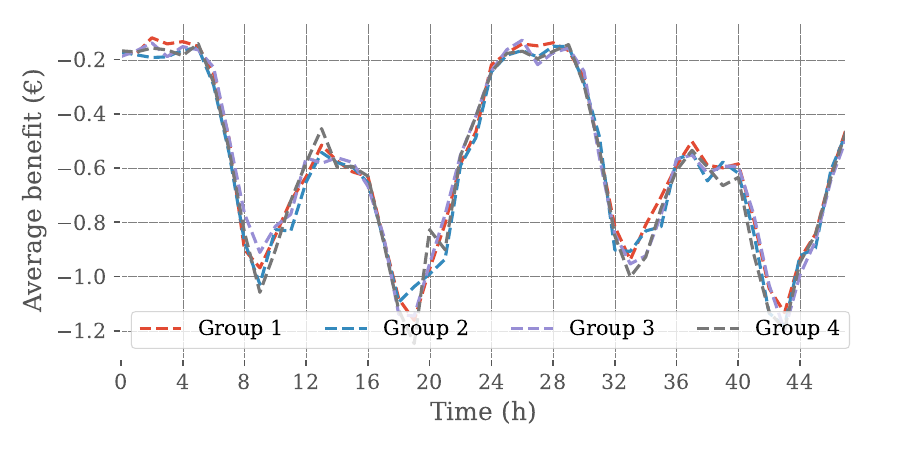} 
        \vspace{-6mm}
        \subcaption{With price-manipulation in LEM B.} \label{fig:prosumer_avg_benefits_4} 
    \end{minipage}
    \caption{Comparison of average benefits for different groups of prosumers under normal and price-manipulated LEM conditions.} \label{fig:prosumer_average_benefits}
    \vspace{-5mm}
\end{figure}

The VAE-GAN model for price manipulation and the MADDPG algorithm for LEM modeling are implemented in Python and run on a Windows machine with an Intel i7 processor.  We summarize the benefits gained by LEM participants from data-driven, model-free decision-making by prosumers under both normal and manipulated pricing scenarios. Prosumers' benefits are computed using $e_i^{d,t}$ and $e_i^{s,t}$, while utility benefits use grid import/export values $e_i^{g,t}$ and $e_i^{g',t}$. The benefit metrics under normal and adversarial pricing are given by:
\begin{subequations}
\begin{align}
& \rho_i^t = \left\{ | e_i^{s,t}| - e_i^{d,t} \right\} \cdot  \lambda^t, && {\text{$\forall i ~\text{and}~ t$}}\\
& \hat{\rho}_i^t =  | e_i^{s,t}| \cdot \tilde{\lambda}^{t,\downarrow}_i - e_i^{d,t} \cdot \tilde{\lambda}^{t,\uparrow}_i, && {\text{$\forall i ~\text{and}~ t$}}. 
\end{align}
\end{subequations}

\vspace{-1mm}
\subsection{Market Performance Without Price Manipulation}
\vspace{-1mm}
\subsubsection{Average Prosumer Benefits Across Market Sizes}
Figs.~\ref{fig:prosumer_avg_benefits_1} and \ref{fig:prosumer_avg_benefits_3} show the average financial benefits for each prosumer group under unmanipulated price conditions for LEM A and LEM B. Across both setups, all groups perform better when prices are unaffected. In LEM A, with fewer prosumers, disparities are more visible. Group 1 consistently records negative benefits due to lacking generation. Groups 2, 3, and 4 engage in both buying and selling, with Groups 3 and 4, having more generation resources, achieving the highest benefits (Table~\ref{tab:total_average_benefit}).

As LEM B scales up, benefit differences between groups shrink. Group 1 still incurs losses, but differences among Groups 2–4 narrow. The larger market results in more balanced trading, reducing individual influence. Our MARL-based framework encourages cooperative behaviors, helping prosumers adapt to market conditions. This stabilization suggests larger LEMs are more resilient to strategic imbalances. While resource-rich prosumers retain their advantage, overall volatility decreases due to collective learning.

\subsubsection{Impact of Battery Capacity on Prosumer Behavior}
Comparing Figs.~\ref{fig:prosumer_bat5kWh_Without} and \ref{fig:prosumer_bat13kWh_Without}, Group 1 remains fully dependent on purchases regardless of battery size. Group 2 shows a slight rise in selling with a 13kWh battery (23.3\% vs. 23.0\%), and a drop in no-operation (4.0\% to 3.3\%), indicating better market engagement. Group 3 shows more notable dependence on battery size: selling rises to 28.2\% (from 26.9\%), and idling drops (6.4\% to 5.1\%). Larger batteries help optimize arbitrage. Group 4 shows minimal change across battery sizes, indicating a well-balanced strategy. Thus, without manipulation, battery capacity impacts behavior in expected ways aligned with storage capabilities.

\vspace{-1.8mm}
\subsection{Market Performance With Price Manipulation}
\vspace{-1.8mm}
\subsubsection{Average Prosumer Benefits Across Market Sizes}
Figs.~\ref{fig:prosumer_avg_benefits_2} and \ref{fig:prosumer_avg_benefits_4} show that adversarial pricing via VAE-GAN leads to reduced financial outcomes for all groups. Manipulated signals distort trade decisions and reduce arbitrage. Table~\ref{tab:total_average_benefit} shows Group 1 suffers the most, with losses deepening from −14.28€ to −28.22€ in LEM B and from −14.08€ to −27.87€ in LEM A. In LEM B, Group 2 shifts from slight profit (0.69€) to loss (−7.51€), while Groups 3 and 4 retain positive but reduced benefits.

\subsubsection{Impact of Battery Capacity on Prosumer Behavior}
Under manipulation, prosumer behavior changes notably. As Fig.~\ref{fig:Distribution_of_actions} shows, Group 1 remains unchanged, limited by lack of assets. In Figs.~\ref{fig:prosumer_bat13kWh_With}-\ref{fig:prosumer_bat5kWh_With}, Group 2 shows adaptive behavior: with 13kWh batteries, selling drops slightly while idling increases to 4.7\%; with 5kWh, idling rises to 7.7\%. Limited storage restricts strategic response under manipulation. Group 3 shows an unexpected increase in selling, possibly due to incomplete adversarial robustness during training. Group 4, with PV and ESS, remains the most stable. With 13kWh, behavior is unchanged; with 5kWh, only slight shifts occur, highlighting the configuration's resilience in adversarial scenarios.

\begin{table*}[t]
\vspace*{1mm} 
\centering
\caption{{Total average benefit of prosumer groups in LEMs over 48 hours with and without price manipulation.}} 
\vspace{-2mm}
\label{tab:total_average_benefit}
\begin{tabular}{
>{\centering\arraybackslash}p{0.12\textwidth} ||
>{\centering\arraybackslash}p{0.18\textwidth}
!{\vrule width 1pt}
>{\centering\arraybackslash}p{0.18\textwidth} ||
>{\centering\arraybackslash}p{0.18\textwidth}
!{\vrule width 1pt}
>{\centering\arraybackslash}p{0.18\textwidth}
}
\hline
\hline
\multirow{4}{*}{\parbox{1.9cm}{\centering \textbf{Prosumer Group} ($I_g$)}} &
\multicolumn{2}{c||}{\textbf{Without manipulation}} &
\multicolumn{2}{c}{\textbf{With manipulation}} \\
& \multicolumn{2}{c||}{$\frac{1}{I_g} \sum_{t \in T} \sum_{i \in I_g} \rho_i^t$} &
\multicolumn{2}{c}{$\frac{1}{I_g} \sum_{t \in T} \sum_{i \in I_g} \hat{\rho}_i^t$} \\
\cline{2-5}
& \textbf{\centering LEM A (\texteuro)} & \textbf{LEM B (\texteuro)} &
  \textbf{LEM A (\texteuro)} & \textbf{LEM B (\texteuro)} \\
\hline
\hline
Group 1 & -14.08 & -14.28 & -27.87 & -28.22 \\
\hline
Group 2 & -5.40 & 0.69 & -13.31 & -7.51 \\
\hline
Group 3 & 3.29 & 8.03 & 1.06 & 7.08 \\
\hline
Group 4 & 3.04 & 8.01 & 0.64 & 6.99 \\
\hline
\hline
\end{tabular}
\vspace{-5mm}
\end{table*}

\begin{figure}[t] 
    \centering
    \begin{minipage}{0.4\textwidth} 
        \centering
        \includegraphics[width=\textwidth]{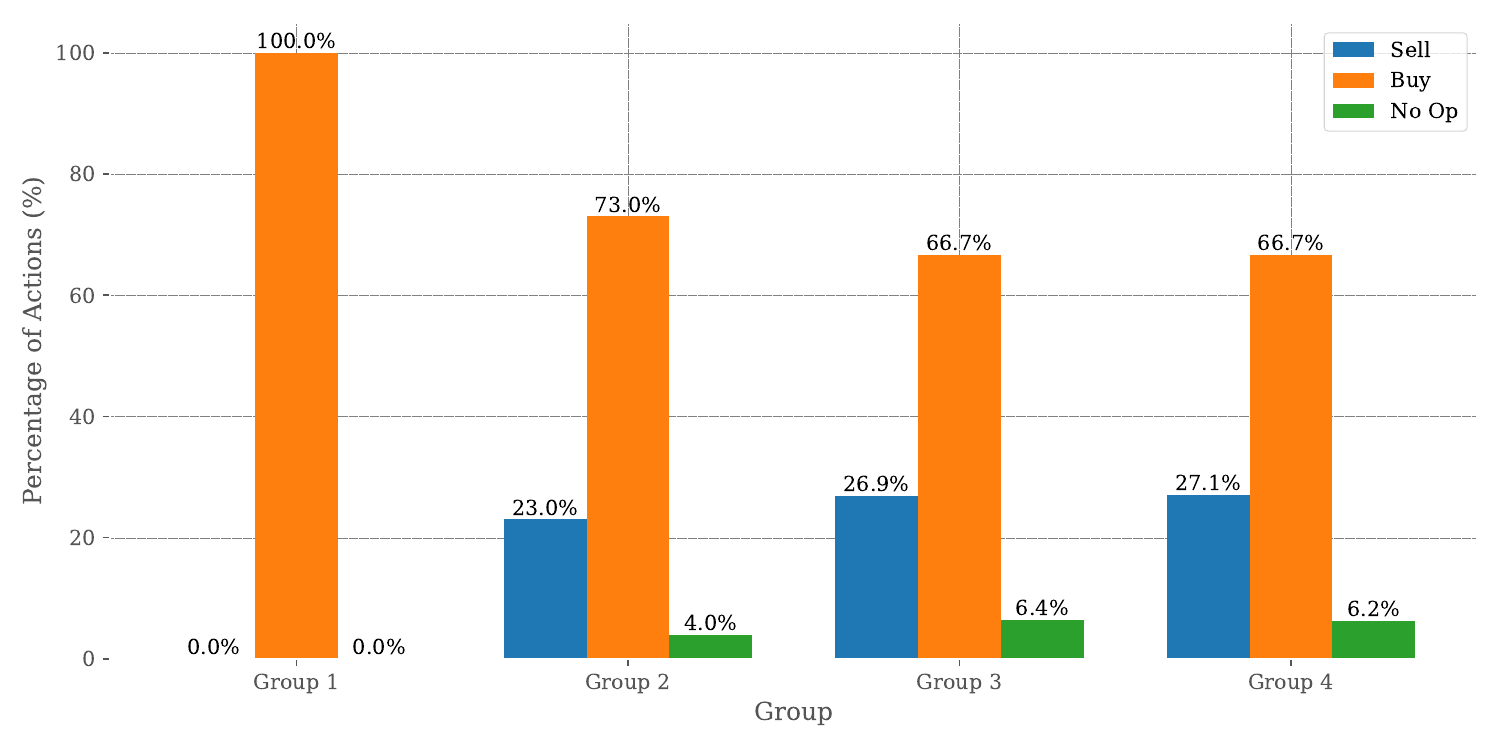} 
        \vspace{-6mm}
        \subcaption{Case 5kWh: Without price-manipulation in LEM B.} \label{fig:prosumer_bat5kWh_Without} 
    \end{minipage}
    \hfill
    \begin{minipage}{0.4\textwidth} 
        \centering
        \includegraphics[width=\textwidth]{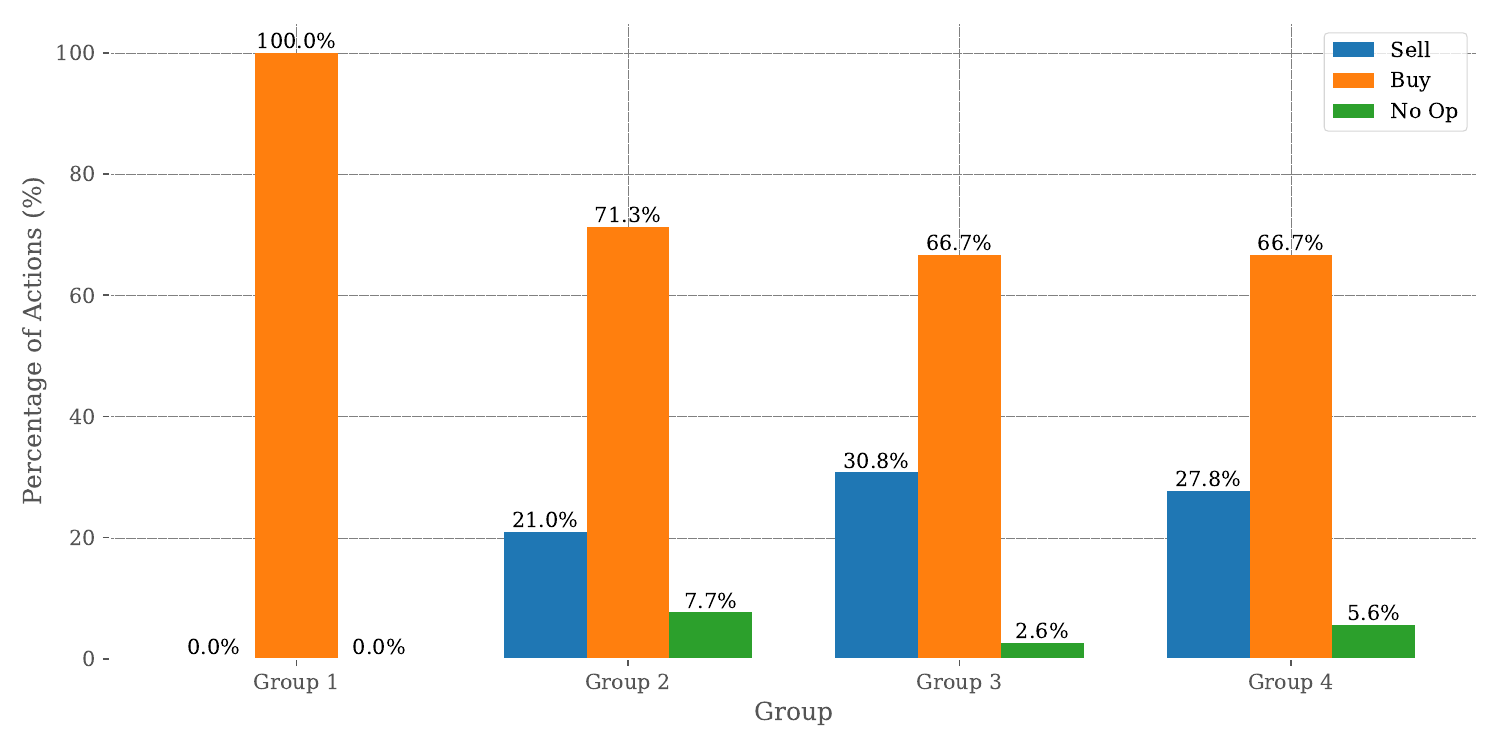} 
        \vspace{-6mm}
        \subcaption{Case 5kWh: With price-manipulation in LEM B.} \label{fig:prosumer_bat5kWh_With} 
    \end{minipage}
    \begin{minipage}{0.4\textwidth} 
        \centering
        \includegraphics[width=\textwidth]{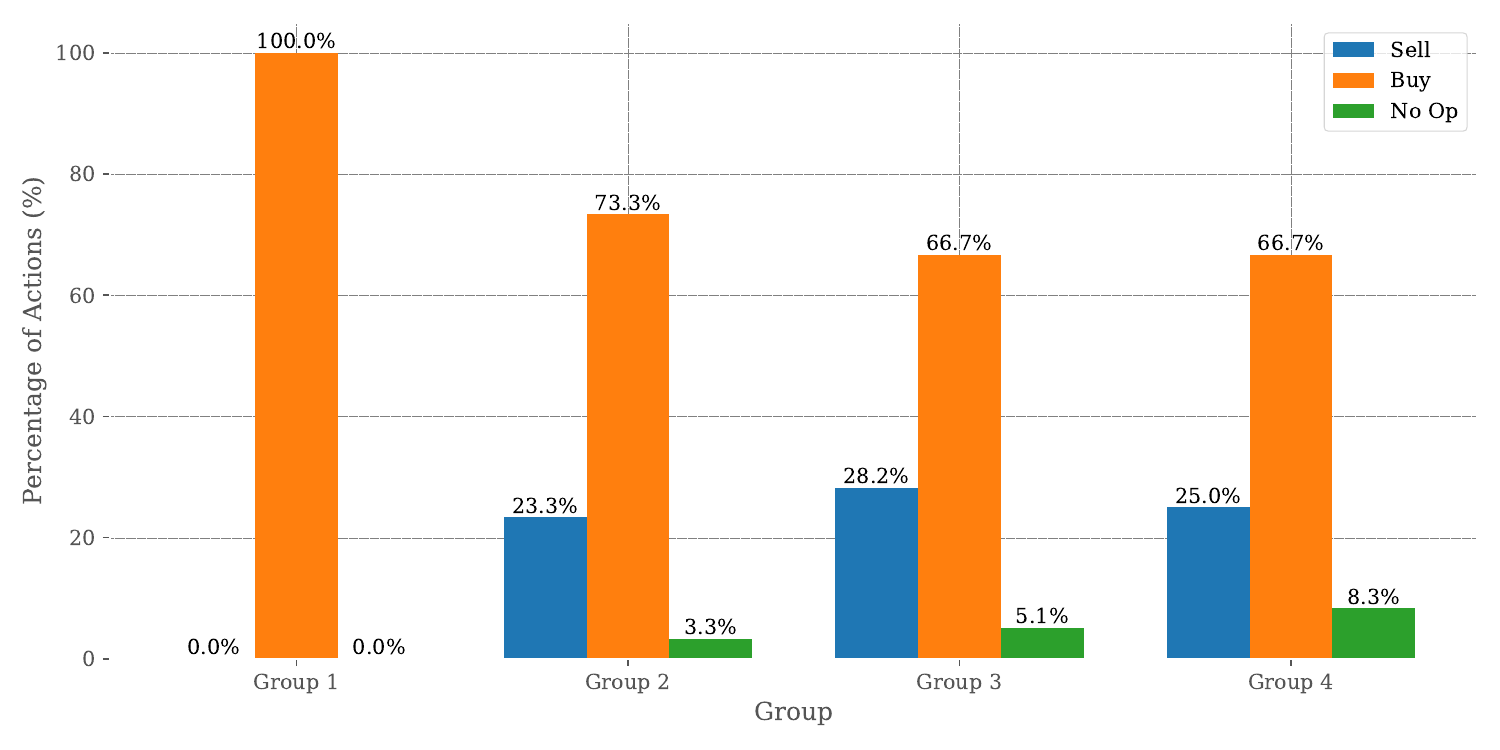} 
        \vspace{-6mm}
        \subcaption{Case 13kWh: Without price-manipulation in LEM B.} \label{fig:prosumer_bat13kWh_Without} 
    \end{minipage}
    \begin{minipage}{0.4\textwidth} 
        \centering
        \includegraphics[width=\textwidth]{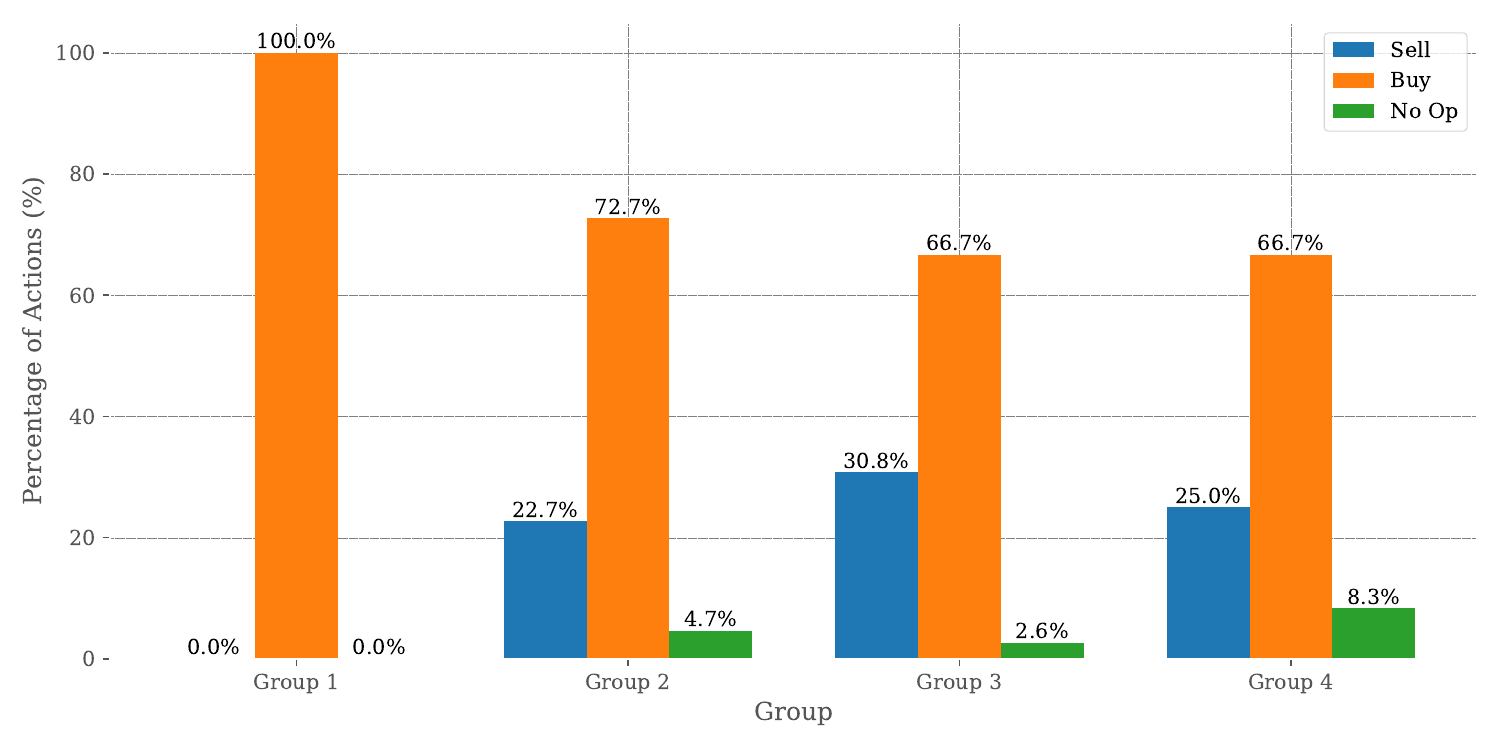} 
        \vspace{-6mm}
        \subcaption{Case 13kWh: With price-manipulation in LEM B.} \label{fig:prosumer_bat13kWh_With} 
    \end{minipage}
    \caption{Distribution of actions percentage among different groups in the large market for various prosumer battery sizes.} \label{fig:Distribution_of_actions}
    \vspace{-5mm}
\end{figure}

\vspace{-3mm}
\section{Conclusion and Future Work}
\vspace{-1mm}

This paper proposes a coordinated LEM framework where heterogeneous prosumer groups interact with a central market entity such as a DSO. To enable decentralized decisions, we adopt a data-driven, model-free MARL approach using the MADDPG algorithm. This actor–critic architecture supports dynamic trading-buying, selling, or idling-based on localized prosumer observations. Our findings show that under normal price conditions, all prosumer groups benefit, with higher returns for those equipped with both generation and storage. As the number of prosumers increases (from LEM A to LEM B), the trading environment stabilizes, reducing group disparities and promoting fairness through emergent cooperation. To explore system vulnerabilities, we introduce a VAE-GAN-based adversarial model that manipulates market prices. This significantly lowers financial benefits, especially for under-resourced prosumers like Group 1, whose losses nearly double. Even Groups 3 and 4 face noticeable declines, particularly in larger markets. Storage-limited agents show increased no-operation behavior, reflecting strategic withdrawal under skewed incentives. Since MADDPG lacks an explicit defense mechanism, prosumers cannot adequately respond, leading to suboptimal outcomes. 

Future work will focus on integrating adaptive strategies into MADDPG to help prosumers counter price manipulation, reduce financial losses, and make informed decisions under adversarial scenarios.

\vspace{-2mm}
\section*{Acknowledgments}
\vspace{-1mm}
This publication is based upon work supported by King Abdullah University of Science and Technology under Award No. RFS-OFP2023-5505.

\vspace{-2mm}

\bibliographystyle{ieeetr}
\bibliography{Ref}

\begin{thebibliography}{10}

\bibitem{pinto2021local}
T.~Pinto, Z.~Vale, and S.~Widergren, {\em Local Electricity Markets}.
\newblock Academic Press, 2021.

\bibitem{TSAOUSOGLOU2022111890}
G.~Tsaousoglou, J.~S. Giraldo, and N.~G. Paterakis, ``Market mechanisms for local electricity markets: A review of models, solution concepts and algorithmic techniques,'' {\em Renew. Sustain. Energy Rev.}, 2022.

\bibitem{morstyn2018using}
T.~Morstyn {\em et~al.}, ``Using peer-to-peer energy-trading platforms to incentivize prosumers to form federated power plants,'' {\em Nature energy}, vol.~3, no.~2, pp.~94--101, 2018.

\bibitem{qiu2021scalable}
D.~Qiu {\em et~al.}, ``Scalable coordinated management of peer-to-peer energy trading: A multi-cluster deep reinforcement learning approach,'' {\em Applied energy}, vol.~292, p.~116940, 2021.

\bibitem{8501599}
L.~Ma {\em et~al.}, ``Real-time rolling horizon energy management for the energy-hub-coordinated prosumer community from a cooperative perspective,'' {\em IEEE Transactions on Power Systems}, vol.~34, no.~2, pp.~1227--1242, 2019.

\bibitem{MAY2023120705}
R.~May and P.~Huang, ``A multi-agent reinforcement learning approach for investigating and optimising peer-to-peer prosumer energy markets,'' {\em Applied Energy}, vol.~334, p.~120705, 2023.

\bibitem{9133518}
H.-M. Chung {\em et~al.}, ``Distributed deep reinforcement learning for intelligent load scheduling in residential smart grids,'' {\em IEEE Trans. on Industrial Informatics}, no.~4, pp.~2752--2763, 2021.

\bibitem{2024_LEMCordWily}
R.~Bian {\em et~al.}, ``A scalable and coordinated energy management for electric vehicles based on multiagent reinforcement learning method,'' {\em Int'l Trans. on Electrical Energy Systems}, vol.~2024, no.~1.

\bibitem{Zhang2015}
G.~Zhang, J.~Lu, and Y.~Gao, {\em Bi-level Programming for Competitive Strategic Bidding Optimization in Electricity Markets}, pp.~315--324.
\newblock Berlin, Heidelberg: Springer Berlin Heidelberg, 2015.

\bibitem{KOCH2019109275}
C.~Koch and L.~Hirth, ``Short-term electricity trading for system balancing: An empirical analysis of the role of intraday trading in balancing {G}ermany's electricity system,'' {\em Renew. Sustain. Energy Rev.}, 2019.

\bibitem{PANOS2019104476}
E.~Panos and M.~Densing, ``The future developments of the electricity prices in view of the implementation of the paris agreements: Will the current trends prevail, or a reversal is ahead?,'' {\em Energy Economics}, vol.~84, p.~104476, 2019.

\bibitem{8957676}
K.~Zhang {\em et~al.}, ``Coordinated market design for peer-to-peer energy trade and ancillary services in distribution grids,'' {\em IEEE Trans. on Smart Grid}, vol.~11, no.~4, pp.~2929--2941, 2020.

\bibitem{10.1007_91}
D.~Palit and N.~Chakraborty, ``Constrained optimal bidding strategy in deregulated electricity market,'' in {\em Artificial Intelligence and Evolutionary Algorithms in Engineering Systems}, pp.~863--873, 2015.

\bibitem{8558521}
T.~Zhao {\em et~al.}, ``Strategic bidding of hybrid ac/dc microgrid embedded energy hubs: A two-stage chance constrained stochastic programming approach,'' {\em IEEE Trans. on Sustainable Energy}, vol.~11, no.~1, pp.~116--125, 2020.

\bibitem{pousinho2015robust}
H.~M. Pousinho {\em et~al.}, ``Robust optimisation for self-scheduling and bidding strategies of hybrid csp–fossil power plants,'' {\em Int'l Journal of Electrical Power \& Energy Systems}, vol.~67, pp.~639--650, 2015.

\bibitem{han2020distributionally}
X.~Han and G.~Hug, ``A distributionally robust bidding strategy for a wind-storage aggregator,'' {\em Electric Power Systems Research}, vol.~189, p.~106745, 2020.

\bibitem{SAMENDE2022119123}
C.~Samende, J.~Cao, and Z.~Fan, ``Multi-agent deep deterministic policy gradient algorithm for peer-to-peer energy trading considering distribution network constraints,'' {\em Applied Energy}, vol.~317, 2022.

\bibitem{8496766}
T.~Chen and W.~Su, ``Local energy trading behavior modeling with deep reinforcement learning,'' {\em IEEE Access}, vol.~6, pp.~62806--62814, 2018.

\bibitem{9705504}
Y.~Ye {\em et~al.}, ``Multi-agent deep reinforcement learning for coordinated energy trading and flexibility services provision in local electricity markets,'' {\em IEEE Trans. on Smart Grid}, 2023.

\bibitem{9351299}
Y.~Deng {\em et~al.}, ``Scenario analysis and anomaly detection of locational marginal price in electricity market,'' in {\em 2020 IEEE Sustainable Power and Energy Conference (iSPEC)}, pp.~2378--2384, 2020.

\bibitem{6074981}
L.~Xie, Y.~Mo, and B.~Sinopoli, ``Integrity data attacks in power market operations,'' {\em IEEE Trans. on Smart Grid}, vol.~2, no.~4, pp.~659--666, 2011.

\bibitem{wang2020market}
X.~Wang and M.~P. Wellman, ``Market manipulation: An adversarial learning framework for detection and evasion,'' in {\em 29th International Joint Conference on Artificial Intelligence}, 2020.

\bibitem{10.1145/3422622}
I.~Goodfellow {\em et~al.}, ``Generative adversarial networks,'' {\em Commun. ACM}, vol.~63, p.~139–144, Oct. 2020.

\bibitem{9578137}
E.~Richardson {\em et~al.}, ``Encoding in style: a stylegan encoder for image-to-image translation,'' in {\em 2021 IEEE/CVF Conference on Computer Vision and Pattern Recognition (CVPR)}, pp.~2287--2296, 2021.

\bibitem{SCHULTZ2024110138}
K.~Schultz {\em et~al.}, ``Convgen: A convex space learning approach for deep-generative oversampling and imbalanced classification of small tabular datasets,'' {\em Pattern Recognition}, vol.~147, p.~110138, 2024.

\bibitem{HSLVAEGAN}
Y.~Liao {\em et~al.}, ``Skip the benchmark: Generating system-level high-level synthesis data using generative machine learning,'' in {\em Proceedings of the Great Lakes Symposium on VLSI 2024}, p.~170–176, 2024.

\bibitem{zhong2020generative}
G.~Zhong {\em et~al.}, ``Generative adversarial networks with decoder--encoder output noises,'' {\em Neural Networks}, vol.~127, pp.~19--28, 2020.

\bibitem{Ratnam14092017}
E.~L. Ratnam {\em et~al.}, ``Residential load and rooftop pv generation: an australian distribution network dataset,'' {\em Int'l Journal of Sustainable Energy}, vol.~36, no.~8, pp.~787--806, 2017.

\bibitem{electricityPrice}
``{DA} electricity prices in {F}rance - {EPEXS}pot.''
\newblock Accessed: 2025-04.

\end{thebibliography}

\end{document}